\documentclass[journal]{IEEEtran}  % Unused option: lettersize
\usepackage{amsmath,amsfonts}
\usepackage{algorithmic}
\usepackage{algorithm}
\usepackage{array}
\usepackage[caption=false,font=normalsize,labelfont=sf,textfont=sf]{subfig}
\usepackage{textcomp}
\usepackage{stfloats}
\usepackage{url}
\usepackage{verbatim}
\usepackage{graphicx}
\usepackage{cite}
\usepackage{hyperref}
\usepackage{tikz}
\usepackage{soul}
\usepackage{xcolor}
\usepackage{mathtools}
\usepackage{adjustbox}
\usepackage{tabularx}
\usepackage{adjustbox}
\usepackage{siunitx}
\usepackage{bm}
\usepackage{upgreek}
\usepackage{multirow}
\usepackage{makecell}
\usepackage{graphicx}
\usepackage{amssymb}
\usepackage{amsthm}
\usepackage{booktabs}
\usepackage{amsmath}
\usepackage[utf8]{inputenc}
\newcommand{\evdata}{$\mathcal{D}_{EV}$}
\newcommand{\evdatacens}{$\mathcal{D}_{EV}'$}

\newcommand{\lbl}{l}
\newcommand{\ourloss}{\mathcal{C}}

\newcommand{\msbest}[2]{$\bm{#1} \hspace{1pt} \pm \hspace{-2.5pt} #2$}
\newcommand{\msequal}[2]{$\mathit{#1} \hspace{1pt} \pm \hspace{-2.5pt} #2$}
\newcommand{\ms}[2]{$#1 \hspace{1pt} \pm \hspace{-2.5pt} #2$}

\DeclareSIUnit{\million}{\text{million}}
\newcommand{\deeplqr}{QNN}
\newcommand{\deeplqrmult}{Multi-QNN}
\newcommand{\deeplinear}{CQNN}

\newcommand{\deepmulti}{Multi-CQNN}

\renewcommand{\vec}[1]{\mathbf{#1}}
\newcommand{\vecbeta}{\boldsymbol{\upbeta}}
\newcommand{\T}{^{\sf T}}
\newcommand{\secref}[1]{Section~\ref{#1}}
\newcommand{\figref}[1]{Fig.~\ref{#1}}
\newcommand{\tabref}[1]{Table~\ref{#1}}
\usepackage{xpatch}
\xpretocmd{\eqref}{Equation~}{}{}

\begin{document}

%\author{IEEE Publication Technology,~\IEEEmembership{Staff,~IEEE,}
        % <-this % stops a space
%\thanks{This paper was produced by the IEEE Publication Technology Group. They are in Piscataway, NJ.}% <-this % stops a space
%\thanks{Manuscript received April 19, 2021; revised August 16, 2021.}}

\title{Modeling Censored Mobility Demand through Censored Quantile Regression Neural Networks}
\author{Frederik Boe Hüttel, Inon Peled, Filipe Rodrigues and Francisco C. Pereira \IEEEmembership{Member, IEEE}
\thanks{Equal contributions between Frederik Boe Hüttel and Inon Peled.}
\thanks{F. B. Hüttel, I. Peled, F. Rodrigues, F. C. Pereira is with the Technical University of Denmark (DTU), Bygning 116B, 2800 Kgs. Lyngby, Denmark. E-mail: fbohy@dtu.dk}
\thanks{July~2022: Accepted for publication in IEEE Transactions on Intelligent Transportation Systems.}
}

%Frederik Boe Hüttel, Inon Peled, Filipe Rodrigues and Francisco C. Pereira}

% \markboth{IEEE Transactions on Intelligent Transportation Systems, Jul~2022}%


\maketitle

\begin{abstract}
    Shared mobility services require accurate demand models for effective service planning.
    On the one hand, modeling the full probability distribution of demand is advantageous because the entire uncertainty structure preserves valuable information for decision-making.
    On the other hand, demand is often observed through the usage of the service itself, so that the observations are \emph{censored}, as they are inherently limited by available supply.
    Since the 1980s, various works on Censored Quantile Regression models have performed well under such conditions.
    Further, in the last two decades, several papers have proposed to implement these models flexibly through Neural Networks. However, the models in current works estimate the quantiles individually, thus incurring a computational overhead and ignoring valuable relationships between the quantiles.
    We address this gap by extending current Censored Quantile Regression models to learn multiple quantiles at once and apply these to synthetic baseline datasets and datasets from two shared mobility providers in the Copenhagen metropolitan area in Denmark. The results show that our extended models yield fewer quantile crossings and less computational overhead without compromising model performance.
\end{abstract}

\begin{IEEEkeywords}
Censored quantile regression, Deep learning, Demand modeling, Latent mobility demand, Multi-task learning, Bayesian modeling.
\end{IEEEkeywords}

\section{Introduction} \label{sec:dcqr-intro}
\IEEEPARstart{S}{hared} mobility services -- e.g., taxis, bike-sharing and ridesourcing -- offer several socio-economic benefits, such as reduced emissions, less traffic congestion and less need for parking~\cite{santos2018sustainability,smith2020towards}.
Effective planning and deployment of such services require reliable estimates of mobility demand, which can be obtained from data-driven modeling~\cite{laporte2018shared,profillidis2019methods}.
The data used in demand modeling is often derived from observations of service usage and are thus inherently limited by available vehicle supply.
Moreover, the data of any mobility service provider does not account for demand lost to competing services and other transport modes.
Consequently, actual demand for mobility is typically latent (i.e., unknown) and its observations are likely to lie below it, namely, they are often \emph{right-censored}.

The \emph{right-censoring} of data makes \textit{uncertainty} quantification essential when modeling demand, to account for the gap between the censored observation and true underlying demand. Uncertainty is critical when balancing the supply to meet the demand, as wrongly estimated demand profiles might lead to insufficient supply. The application of censored modeling and supply balancing for bike-sharing services are analyzed in~\cite{gamelli2022predictive}.

Previously, Gaussian Processes have been the go-to model for modeling censored mobility demand which yields a complete distribution of latent demand~\cite{gammelli2020estimating}.
However, while Gaussian Processes allow for a flexible, non-parametric fit, they still impose a Gaussian assumption on the latent distribution and face limitations when scaling to large datasets. The latter issue is becoming increasingly influential in the transportation domain and research as large datasets are increasingly used in transportation modeling~\cite{liu2021deepTSP, li2020short}, requiring models to scale seamlessly for adequate censored modeling in the transportation domain.

This work proposes to model latent mobility demand via Multi-Output Censored Quantile Regression Neural Networks (Multi-CQNN).
These models do not face the limitations mentioned above as they make no assumptions on the parametric form of the latent distribution and scale well to large datasets.
As their name implies, Multi-CQNN estimates multiple quantiles of the predictive distribution while accounting for censorship in the observed demand.
By being multi-output, they also address two drawbacks of single-output CQNN models, where estimating quantiles individually both incurs a significant computational overhead and yields crossing quantiles, wherein a lower quantile crosses a higher one~\cite{koenker1984l-estimation}. We demonstrate the advantages of Multi-CQNN empirically on synthetic data as well as real-world shared mobility data.

To the best of our knowledge, this is the first work to apply (Multi-)CQNN in the transport domain.
Furthermore, whereas existing works on CQNN often assume fixed censorship thresholds, we experiment with dynamic and random censorship thresholds, which make for a more complex modeling setting.
We also provide a Python implementation of Multi-CQNN in \url{https://github.com/inon-peled/CQRNN-pub/}.

The rest of this work is organized as follows.
In~\secref{sec:dcqr-litrev}, we review works related to CQNN and identify knowledge gaps, particularly in the transport domain.
\secref{sec:dcqr-mth} then describes our Multi-CQNN methodology.
Next, we demonstrate the advantages of our methodology via experiments: first with synthetic censored data in~\secref{sec:dcqr-exp-synth}, then with real-world censored data from two shared mobility services (bikes and electric cars) in~\secref{sec:dcqr-exp-real}.
Finally,~\secref{sec:dcqr-conclusion} summarizes our findings and outlines our future work plans.

\section{Literature Review} \label{sec:dcqr-litrev}
In this Section, we review existing works on non-censored and censored Quantile Regression (QR), and in particular, Neural Network-based QR. We focus our review on censored QR approaches, as there also exists plenty of work other than QR to model censored data. For example,~\cite{robert2009impacts} and \cite{kim2017systematic} systematically analyze and study the difference between point and section data from different sensors, to predict travel speeds and the effect on mobility service.

For the more general topics of Censored Regression and Neural Networks, we kindly refer the reader to the following resources.
A review of censored and non-censored methods for mobility demand modeling appears in our recent joint work with~\cite{gammelli2020estimating}.
The general theory and practice of Neural Networks is well studied in~\cite{nielsen2015neural} and~\cite{Goodfellow-et-al-2016}, and several of their recent applications in the transport domain are reviewed in~\cite{himanen2019neural} and \cite{liu2021deepTSP}.
Let us first introduce the general method of Quantile Regression~\cite{koenker1978regression}, regardless of censorship.

For any probability distribution and $0 < \theta < 1$, the $\theta$'th~\emph{quantile} is the smallest value at which the cumulative probability mass is $\theta$.
QR approximates a latent distribution by estimating its quantiles for given $\theta$'s, and thus does not presume any parametric form for the distribution.
The regression itself can follow any functional form -- e.g., linear~\cite{koenker1978regression}, nonlinear~\cite{koenker1996interior}, multivariate~\cite{carlier2016vector} or nonparametric~\cite{zheng2012qboost} --
and the quantiles can be combined into a fully estimated distribution~\cite{quinonero2006evaluating, cannon2011quantile}. 

Importantly, the fully estimated distribution preserves useful information, which might otherwise be lost through the more common practice of estimating only a few central moments, such as mean and standard deviation~\cite{peled2019preserving}.
In turn, the preserved information allows for better informed decisions, e.g., service operators can use the full uncertainty structure of future demand to decide whether to balance the fleet conservatively or more opportunistically.
In addition, by taking values of $\theta$ close to $0$ and $1$, QR can be more robust to outliers than mean regression~\cite{koenker2005quantile}. 

A variant of QR is Censored Quantile Regression (CQR), where observed realisations from a latent distribution are assumed to be clipped at some thresholds.
The distribution of these observations is then a mixture of continuous variables (within the observable range) and discrete variables (at each threshold).
Many works on CQR build upon the early formulation by~\cite{powell1984least, powell1986censored}, as we ultimately do too in~\secref{sec:dcqr-mth}.
Some of these works focus on estimators of derived CQR formulations~\cite{buchinsky1998recent, leng2013quantile, galvao2013estimation, cheng2014quantile, chernozhukov2015quantile}.
Other works on CQR develop more complex and non-parametric models, as we next review, starting with models that are not based on Neural Networks.

\cite{yu2007bayesian} devise a Bayesian Inference approach to linear CQR, in which they use the Asymmetric Laplace likelihood, as also common in other Bayesian CQR works~\cite{yang2016posterior}.
They evaluate their approach on some commonly used, censored data baselines: synthetic datasets, which we too use in~\secref{sec:dcqr-exp-synth}, and a real-world dataset of women's labor force participation~\cite{mroz1987sensitivity}.
\cite{gannoun2005non} develop a CQR model based on local linear approximations and evaluate it on synthetically generated datasets.
\cite{shim2009support} offer a Support Vector Machine-based CQR model, which they evaluate on the same synthetic datasets as in~\cite{gannoun2005non}, and on a real-world dataset of heart transplant survival. 
\cite{li2020censored} propose a Random Forest-based CQR, which they compare with other Random Forest models on both synthetic data and two real-world datasets: housing prices and a biliary disease.

In the last two decades, Neural Networks have increasingly been used for Quantile Regression (QNN) in multiple research areas, taking advantage of their flexible nonlinear modeling capabilities.
For non-censored regression,~\cite{taylor2000quantile} provides an early form of QNN with a single dense hidden layer, and uses it to estimate a latent distribution of multiperiod financial returns.
In studies of the electric power industry, He et al. use non-censored QNN to estimate latent distributions of electricity production~\cite{he2018probability, he2020probability} and consumption~\cite{he2019electricity}, while~\cite{haben2016hybrid} and~\cite{zhang2018improved} use non-censored QNN to predict electricity loads.
In the transport domain,~\cite{tian2020probabilistic} use a non-censored QNN with a single hidden neuron to predict \SI{15}{\min} air traffic in a Chinese airport, and~\cite{rodrigues2020beyond} devise a non-censored, multi-output QNN that jointly estimates mean and quantiles, whereby they predict taxi demand in New York City, in 30 minutes intervals.

The multi-output QNN reduces the computational overhead of estimating multiple quantiles independently as the full latent distribution is estimated in one forward pass. Traditionally QR regression models face the problem of quantile crossing, wherein a lower quantile function crosses a higher one~\cite{koenker1984l-estimation}. While this problem has been well studied for non-censored cases~\cite{He1997quantile,takeuchi2006nonparametric, sangnier2016joint, chernozkukov2010quantiles, rodrigues2020beyond,park2021learning}, our work is the first to study it in the context of censored data.

In fact, very few works apply QNN in a Censored setting (CQNN).
\cite{cannon2011quantile} develops a general architecture for both QNN and CQNN, which he implements as an \texttt{R} package.
He uses a smoothing technique by~\cite{chen2007finite} to replace the loss function with a differentiable approximation, which is amenable to gradient-based training, and applies the implementation in a censored case study of precipitation forecasting.
\cite{jia2020deep} propose another CQNN model, similar to but with deeper architectures than that of~\cite{cannon2011quantile}.
They implement their model in Python via \texttt{Keras} and apply it to censored survival datasets: a synthetic dataset and a breast cancer dataset.

In conclusion,~\cite{cannon2011quantile} and~\cite{jia2020deep} are yet the only studies on Quantile Regression Neural Networks in a Censored setting.
In particular, there are no works on alleviating the computational cost of CQNN or the quantile crossing problem for censored cased studies. Moreover, there are no works on CQNN in the transport domain, despite the prevalent censorship in transport data with complex network-structures, behavioural feedback and demand and supply dynamics~(\secref{sec:dcqr-intro}). 
We address all these gaps by devising a Multi-Output Censored Quantile Regression Neural Network and applying it to several datasets of real-world shared mobility services.

\section{Methodology} \label{sec:dcqr-mth}

A censored dataset consists of covariates $\vec{x}_1, \dots, \vec{x}_N$ and corresponding observations $y_1, \dots, y_N$, which are clipped versions of latent variables $y^*_i, \dots, y^*_N$.
Namely, for some thresholds $\tau_1, \dots, \tau_N$, all the observations are either left-censored or right-censored, such that:
\begin{align}
    y_i &= 
    \begin{cases}
    y^* \,, & y^*_i > \tau_i \\
    \tau_i \,, & y^*_i \leq \tau_i
    \end{cases}
    \text{ in left-censorship,}
    \\
    y_i &= 
    \begin{cases}
    y^* \,, & y^*_i < \tau_i \\
    \tau_i \,, & y^*_i \geq \tau_i
    \end{cases}
    \text{ in right-censorship.}
\end{align}
Each threshold is either given or unknown, and if censorship is fixed, then $\tau_1 = \cdots = \tau_N$.
$y^*_1, \dots, y^*_N$ are drawn from a latent distribution, whose $\theta$'th quantiles we wish to estimate for some $0 < \theta < 1$. 
For a quantile regression model with parameters $\vecbeta$, $y_i=\max(0,y^*_i)$ (i.e., left-censorship at 0) and a specified quantile $\theta$, the following likelihood function can be used for Bayesian Inference~\cite{yu2007bayesian}:

%\begin{equation}
    %\mathcal{C}\left(y_1\,, \dots \,, y_N \Big\vert \vecbeta, \vec{x}, \theta \right)
    %=
    %\theta^N \left(1 - \theta\right)^N \exp\Bigg\{ -\sum_{i=1}^N \rho_\theta \left( y_i - \max \left\{ 0, \hat{q}_{i,\theta}\right\} \right) \Bigg\}
%    \,, \label{eq:yslik}
%\end{equation}

\begin{multline}
    \mathcal{C}\left(y_1\,, \dots \,, y_N \Big\vert \vecbeta, \vec{x}, \theta \right)
    = \theta^N \left(1 - \theta\right)^N \\ 
    \exp\Bigg\{ -\sum_{i=1}^N \rho_\theta \left( y_i - \max \left\{ 0, \hat{q}_{i,\theta}\right\} \right) \Bigg\}
    \,, \label{eq:yslik}
\end{multline}

where $\hat{q}_{i,\theta}$ is the estimated $\theta$'th quantile of $y_i$ and $\rho: \mathbb{R} \rightarrow \mathbb{R}$ is the Tilted Loss (TL) function,
\begin{equation}
    \label{eq:tl}
    \rho_\theta(r) = \max\{\theta r\,, (\theta - 1)r \}
    \,.
\end{equation}

\begin{figure}[tb]
    \centering
    \includegraphics[width=0.49\textwidth]{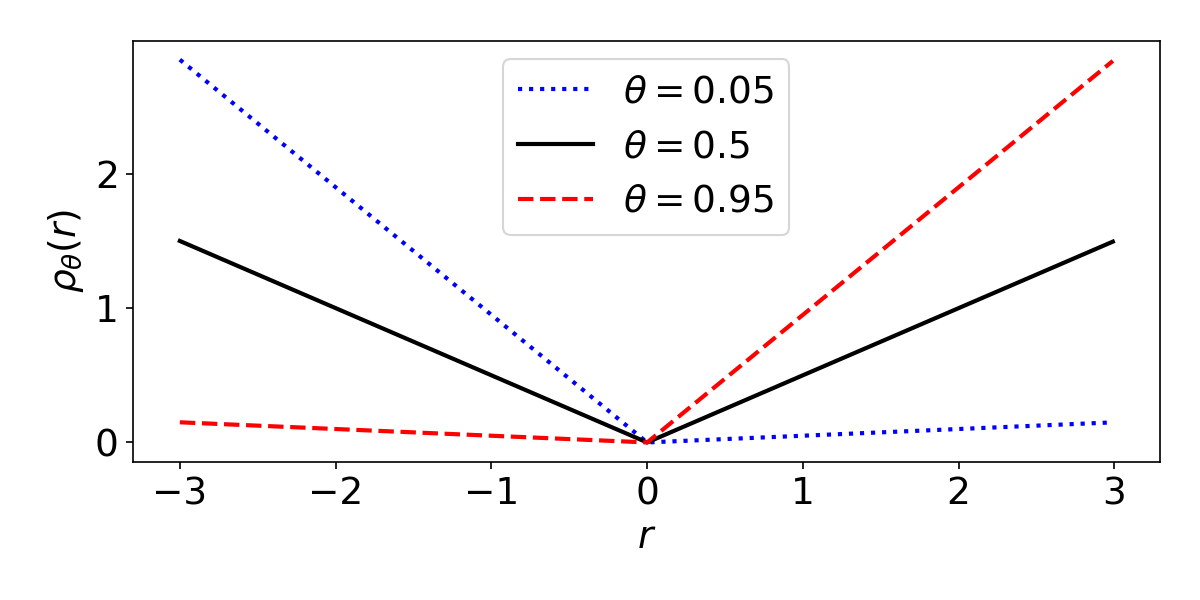}
    \caption{Tilted Loss.}
    \label{fig:tiltloss}
\end{figure}

\figref{fig:tiltloss} illustrates how TL penalizes the prediction error $r = \hat{q}_{i,\theta} - y_i$ in a manner that depends on $\theta$.
For the median ($\theta = 0.5$), the loss is the same regardless of the sign of $r$.
For quantiles above the median (e.g., $\theta = 0.95$), the loss is worse for $y_i > \hat{q}_{i,\theta}$ than for $y_i < \hat{q}_{i,\theta}$ with the same magnitude of $r$, and vice versa for quantiles below the median (e.g., $\theta = 0.05$).
For any $\theta$, the loss equals zero if $y_i = \hat{q}_{i,\theta}$ and is otherwise positive.

Based on \eqref{eq:yslik}, we get the following likelihood function for left-censorship at a stochastic threshold $\tau_i$:

\begin{multline}
\ourloss\left(y_1 \,, \dots \,, y_N \Big\vert \vecbeta, \vec{x}, \theta \right)
    =
    \theta^N \left(1 - \theta\right)^N 
     \\ 
    \exp\Bigg\{ -\sum_{i=1}^N \rho_\theta \left( y_i - \max \left\{ {\color{black} \tau_i},  {\color{black} \hat{q}_{i,\theta}}\right\} \right) \Bigg\}
    \,.
    \label{eq:ourlik}
\end{multline}
Note that $\tau_i$ must be specified also for all observations (both censored and non-censored).

\def\layersep{2.5cm}
\subsection{Multi-Output Censored Quantile Regression Neural Network}
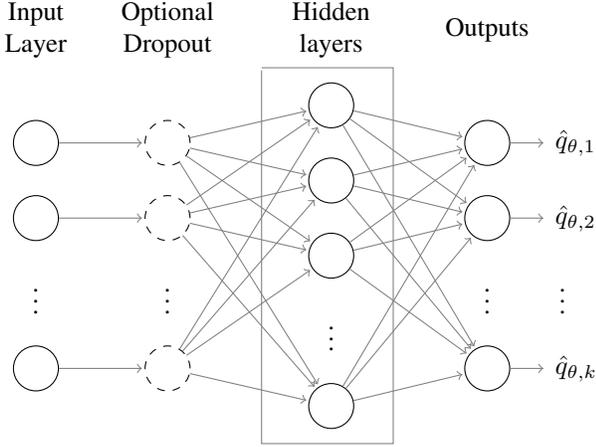
\begin{figure}[tb]
    \centering
    \begin{tikzpicture}[node distance=\layersep,shorten >=1pt,->,draw=black!50,] 
    \tikzstyle{vdots}=[];
    %\tikzstyle{every pin distance = \layersep};
    \tikzstyle{every pin edge}=[<-,shorten <= 0pt];
    \tikzstyle{neuron}=[circle,minimum size=17pt,inner sep=0pt];
    \tikzstyle{negate neuron}=[neuron,draw=black];
    \tikzstyle{dropout neuron}=[neuron,draw=black,dashed];
    \tikzstyle{input neuron}=[neuron,draw=black];
    \tikzstyle{output neuron}=[neuron,draw=black];
    \tikzstyle{hidden neuron}=[neuron,draw=black];
    \tikzstyle{annot} = [text width=4em, text centered]

    % Inputs
    \node[vdots] (I-3) at (-1.25,-3) {$\vdots$};
    \node[input neuron] (I-1) at (-1.25,-1) {};
    \node[input neuron] (I-2) at (-1.25,-2) {};
    \node[input neuron] (I-4) at (-1.25,-4) {};

    % Negate input
    %\node[vdots] (NI-3) at (-1.25,-3) {$\vdots$};
    %\node[input neuron] (NI-1) at (-1.25,-1) {$-$};
    %\node[input neuron] (NI-2) at (-1.25,-2) {$-$};
    %\node[input neuron] (NI-4) at (-1.25,-4) {$-$};

    % Draw the Dropout layer nodes
    \node[vdots] (D-3) at (0.5,-3) {$\vdots$};
    \foreach \name / \y in {1,2,4}
    % This is the same as writing \foreach \name / \y in {1/1,2/2,3/3,4/4}
        \node[dropout neuron] (D-\name) at (0.5,-\y) {};

    % Draw the hidden layer nodes
    \draw[style=-] (1.75,0) -- (3.5,0) -- (3.5,-5) -- (1.75,-5) -- (1.75,0);
     \path[yshift=0.5cm]
            node[vdots] (H-4) at (\layersep + 5,-4 cm) {$\vdots$};
    \foreach \name / \y in {1,2,3,5}
        \path[yshift=0.5cm]
            node[hidden neuron] (H-\name) at (\layersep + 5,-\y cm) {};
            
    % Negate output
    %\node[output neuron, right of=H-3,xshift=-0.5cm] (O) {};
    %\node[negate neuron,right of=O,pin={[pin %edge={->}]right:$\hat{q}_\theta$},xshift=-0.9cm] (NO) {$-$};
    %\draw (O) edge (NO);
    
    \node[vdots] (O-3) at (4.75,-3) {$\vdots$};
    \node[vdots] (t) at (5.75,-3) {$\vdots$};
    %\node[output neuron, right of=H-3,xshift=-0.5cm] (O-1) {};
    %\node[output neuron, right of=H-3,xshift=-0.5cm] (O-2) {};
    %\node[output neuron, right of=H-3,xshift=-0.5cm] (O-4) {};
    \node[output neuron,pin={[pin edge={->}]right:$\hat{q}_{\theta,1}$}] (O-1) at (4.75,-1) {};
    \node[output neuron,pin={[pin edge={->}]right:$\hat{q}_{\theta,2}$}] (O-2) at (4.75,-2) {};
    \node[output neuron,pin={[pin edge={->}]right:$\hat{q}_{\theta,k}$}] (O-4) at (4.75,-4) {};
    %\node[input neuron] (I-1) at (-3,-1) {};
    %\node[input neuron] (I-2) at (-3,-2) {};
    %\node[input neuron] (I-4) at (-3,-4) {};

    % Connect nodes.
    %\foreach \source in {1,2,4}
        %\draw (I-\source) edge (NI-\source);
    \foreach \source in {1,2,4}
        \draw (I-\source) edge (D-\source);

    % Connect every node in the dropout layer with every node in the
    % hidden layer.
    \foreach \source in {1,2,4}
        \foreach \dest in {1,2,3,5}
            \path (D-\source) edge (H-\dest);
    % Connect every node in the hidden layer with the output layer
    %\foreach \source in {1,2,3,5}
    %    \path (H-\source) edge (O);
    \foreach \source in {1,2,4}
        \foreach \dest in {1,2,3,5}
            \path (H-\dest) edge (O-\source);

    % Annotate the layers
    %\node[annot,above of=NI-1, node distance=1.5cm] {Negate};
    \node[annot,above of=I-1, node distance=1.5cm] {Input Layer};
    \node[annot,above of=H-1, node distance=1cm] (hl) {Hidden layers};
    \node[annot,above of=D-1, node distance=1.5cm] (dl) {Optional Dropout};
    \node[annot,above of=O-1, node distance=1.5cm] (ol) {Outputs};
    %\node[annot,right of=ol,xshift=-0.9cm] {Negate};
    \end{tikzpicture}
    \caption{Our Multi-Output Censored Quantile Regression Neural Network architecture.}
    \label{fig:Mdcqr-arch}
\end{figure}

A naive approach to QNN is to independently fit a Neural Network (NN) for each value of $\theta$, while using either \eqref{eq:yslik} or (\eqref{eq:ourlik}) as a loss function.
As the computational cost of training multiple NNs can be high, we propose instead to use a Multi-Output Censored Quantile Regression Neural Network (Multi-CQNN). 
The corresponding architecture models multiple quantiles simultaneously, so that its output dimensionality equals the number of desired quantiles.
That is, the Multi-CQNN architecture has an output neuron for each of the $K$ different quantiles $\{\theta_k\}_{k=1}^K$, as depicted in \figref{fig:Mdcqr-arch}. By estimating all $K$ quantiles together in one forward pass, Multi-CQNN eliminates the computational cost of independent training  without drastically increasing the number of trainable parameters, as these are shared in the NN layers.
Multi-CQNN can be viewed as a multi-task learner~\cite{ruder2017multi}, where each output node has the task of estimating the quantile related to that node. We extend the loss from \eqref{eq:ourlik} to the multi-output case by summing the loss from each task:
\begin{multline}
\ourloss\left(y_1 \,, \dots \,, y_N \Big\vert \vecbeta, \vec{x}, \{\theta_k\}_{k=1}^K \right)
    = \sum_{k=1}^K \Bigg(
    \theta_k^N \left(1 - \theta_k\right)^N \\
    \exp\Bigg\{ -\sum_{i=1}^N \rho_{\theta,k} \left( y_{i} - \max \left\{ {\color{black} \tau_i},  {\color{black} \hat{q}_{\theta,k, i}}\right\} \right) \Bigg\}
    \,\Bigg),
    \label{eq:multiourlik}
\end{multline}

where $\hat{q}_{i,\theta,k}$ is the NN ouput for quantile $k$.
If $K=1$, so that only a single quantile is estimated, then ~\eqref{eq:multiourlik} indeed reduces to ~\eqref{eq:ourlik}. When $K > 1$, the NN parameters are shared across the different quantiles, and this has a regularising effect on the parameters and outputs.

In particular, this property is effective in alleviating quantile crossing, as we later show in \secref{sec:dcqr-exp-crossing}.
As noted in ~\cite{He1997quantile}, quantile crossing is primarily caused by estimating quantiles individually, and so can be alleviated by limiting the flexibility in individual estimation~\cite{rodrigues2020beyond}. Parameter sharing indeed limits this flexibility in Multi-CQNN by forcing it to learn a latent data representation that accounts for multiple quantiles at once.

\subsection{Optimization}
When dealing with right-censored datasets in next Sections, we slightly modify the architecture by negating the output quantiles and mirroring them (e.g., swapping the output for $\theta=0.05$ with the output for $\theta=0.95$). In this manner, the NN treats right-censored data as if it were left-censored.
We fit the parameters of the models~(denoted $\vecbeta$) by minimizing the negative log-likelihood of \eqref{eq:ourlik} and (\ref{eq:multiourlik}) using backpropagation with the Adam optimizer~\cite{kingma2014adam}.
Minimisation of the log-likelihood functions mentioned above simplifies to minimisation of the censored quantile error function~\cite{cannon2011quantile, koenker2008censored}: 

\begin{equation}
    \mathcal{L}(\vecbeta)=\sum_{i=1}^N\rho_\theta \left( y_{i} - \max \left\{ {\color{black} \tau_i},  {\color{black} \hat{q}_{i,\theta}}\right\} \right)
\end{equation}
and for the multi-output case:
\begin{equation}
    \mathcal{L}(\vecbeta)=\sum_{k=1}^K\sum_{i=1}^N\rho_{\theta,k} \left( y_{i} - \max \left\{ {\color{black} \tau_i},  {\color{black} \hat{q}_{\theta,k, i}}\right\} \right)
\end{equation}
We use a learning rate of $0.01$ with norm clipping at $1$ and $\ell_2$ regularisation with $\lambda=0.001$, while using a validation set for early stopping.
Additional details about architecture, initialisation, fitting and evaluation depend on the dataset and manner of experimentation, as described in the following Sections.

\section{Experiments for Demonstrating the Advantages of Censored Quantile Regression} \label{sec:dcqr-exp-synth}

In this Section, we empirically demonstrate some advantages of using a Multi Output Censored Quantile Regression Neural Networks (Multi-CQNN).
First, we compare censorship-aware with censorship-unaware Quantile Regression models and show that censorship-awareness can better reconstruct latent values for both single- and multi-output models.
Then, we compare Multi-CQNN to parametric Censored Regression and show the advantages and disadvantages of each modeling method. We then compare the quantile crossing problem between the single and Multi-CQNN, and show that our model has substantially less quantile crossings compared with the single output model, both for the censored and non-censored observations.

The experiments in this Section are based on commonly used, synthetic baseline datasets, and the subsequent Section proceeds to apply Multi-CQNN to model real-world transportation datasets. Each experiment is performed $10$ times with equal starting conditions for all models. In the following tables, we report the average over these $10$ runs and measure uncertainty via their standard deviation.

\subsection{Non-censored vs. Censored Quantile Regression} \label{sec:dcqr-exp-nocens}

Let us first show that the predictive quality of Quantile Regression Neural Networks can improve by accounting for data censorship.
We use the same synthetic baseline datasets as in~\cite{yu2007bayesian}, as they consider a censored parametric Bayesian model, where the latent variable is

\begin{equation}
    y^* = x_0 + x_1 + x_2 + \varepsilon
    \,,
\end{equation}
where $x_0 = 1 \,, x_1 \in \{-1, 1\} \,, x_2 \in \mathbb{R}$ and the noise $\varepsilon$ follows some distribution with $0$ mean.
Left-censorship occurs at zero, so that we observe
\begin{equation}
    y = \max\{0, y^*\} \,.
\end{equation}

For any random variable $A$ and $0\!<\! \theta\!<\!1$, let $q_\theta(A | \vec{x})$ denote the $\theta$'th conditional quantile of $A$ given $\vec{x} = \left[x_0, x_1, x_2\right]\T$.
Hence:
\begin{equation}
    q_\theta(y | \vec{x}) = \max\{0 \,, q_\theta(y^* | \vec{x})\} = \max\{0 \,, x_0 + x_1 + x_2 + q_\theta(\varepsilon | \vec{x})\}
    \,.
    \label{eq:qcond}
\end{equation}

\begin{figure*}[tb]
    \centering
    \includegraphics[width=\textwidth]{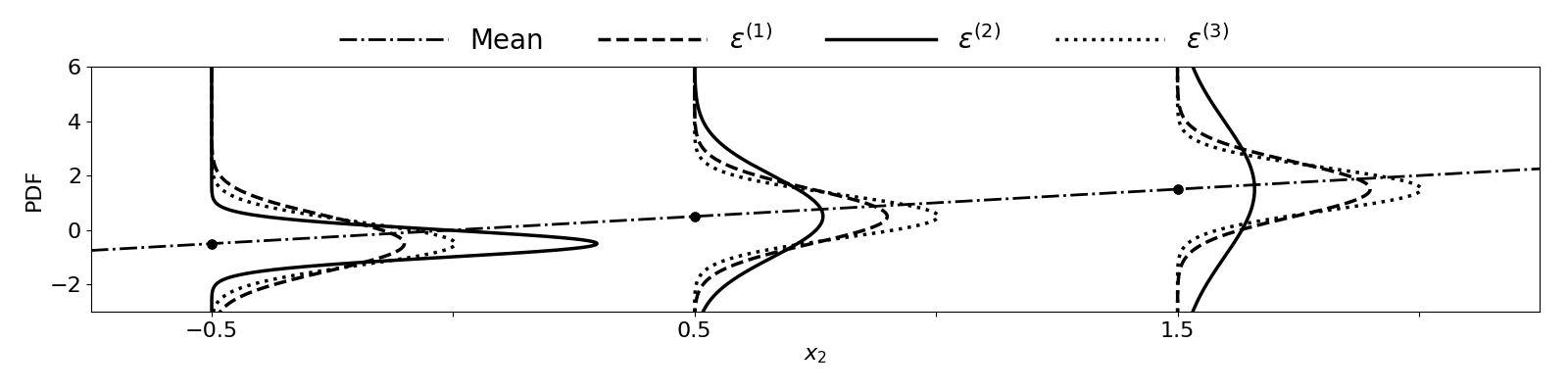}
    \caption[Synthetic $y^*$ distributions]{
    Distribution of synthetic $y^*$ for $x_0 = 1, x_1 = -1, x_2 = -0.5, 0.5, 1.5$ and each noise distribution $\varepsilon^{(j)}$.
    In each case, $y^*$ follows a Gaussian distribution with mean $x_0 + x_1 + x_2$ (dash-dotted line).
    For the Standard Gaussian noise (dashed) and Gaussian Mixture noise (solid), the distribution of $y^*$ is homoskedastic, i.e., has fixed variance $1$ and $0.791^2$, respectively.
    For the Heteroskedastic noise (dotted), $y^*$ has variance that changes with $x_2$ as $(1 + x_2)^2$.
    }
    \label{fig:synthds}
\end{figure*}

\begin{figure*}[b]
    \centering
    \includegraphics[width=0.7\textwidth]{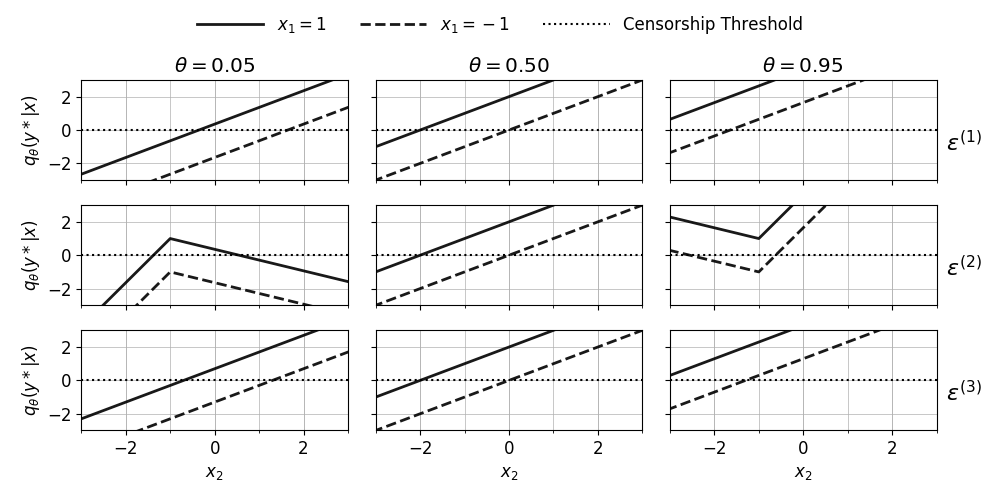}
    \caption[Conditional quantiles of synthetic $y^*$]{Conditional quantiles of synthetic $y^*$ for each noise distribution}
    \label{fig:qsynth}
\end{figure*}

Similarly to \cite{yu2007bayesian}, we experiment with $\theta = 0.05, 0.50, 0.95$ and three noise distributions,
\begin{align}
    \text{Standard Gaussian:} \qquad & \varepsilon^{(1)} \sim \mathcal{N}(0, 1)
    \,, \\
    \text{Heteroskedastic:} \qquad & \varepsilon^{(2)} \sim (1 + x_2)\mathcal{N}(0, 1)
    \,, \\
    \text{Gaussian Mixture:} \qquad & \varepsilon^{(3)} \sim 0.75\mathcal{N}(0, 1) + 0.25\mathcal{N}(0, 2^2)
    \,.
\end{align}
The corresponding conditional quantiles of $y$ are thus 
\begin{equation}
q_\theta\left(y^{(j)} | \vec{x}\right) = \max \left\{ 0 \,, q_\theta\left(y^{*\,,j} | \vec{x}\right) \right\}
\,,
\end{equation}
such that:
\begin{align}
    q_\theta(y^{*\,,1} | \vec{x}) &= \Phi^{-1}\Big(x_0 + x_1 + x_2 \,, 1 \Big) \,, \label{eq:q1} \\
    q_\theta(y^{*\,,2} | \vec{x}) &= \Phi^{-1}\Big(x_0 + x_1 + x_2 \,, \left| 1 + x_2 \right| \Big) \,, \label{eq:q2} \\
    q_\theta(y^{*\,,3} | \vec{x}) &= \Phi^{-1}\Big(x_0 + x_1 + x_2 \,, \sqrt{0.75^2 + 0.25^2} \Big) \,, \label{eq:q3} 
\end{align}
where $\Phi^{-1}\left(\mu, \sigma\right) : \left[0 \,, 1 \right] \rightarrow \mathbb{R}$ is the quantile function of $\mathcal{N}(\mu, \sigma^2)$.
\figref{fig:synthds} illustrates the distribution of $y^*$ with each noise, for $x_1 = 1$ and several values of $x_2$.
\figref{fig:qsynth} illustrates the conditional quantiles of $y^*$ for each noise and $\theta$.
For Heteroskedastic noise, the conditional quantiles of $y^*$ are non-linear, as their slopes change at $x_2 = -1$.

For each $\varepsilon^{(j)}$, $j=1,2,3$, we generate a synthetic dataset by independently drawing $N = 1000$ samples from $\varepsilon^{(j)}$ and $\vec{x}$, where
\begin{align}
    x_0  = 1 
    \,, \qquad
    x_1 \sim \mathcal{U}\{-1, 1\} 
    \,, \qquad
    x_2 \sim \mathcal{N}\left(0, 1\right)
    \,.
\end{align}
We then also compute the corresponding $y^{*}$ and $y$, and obtain that approximately $30\%$ of the observations $y_1^{(j)}, \dots, y_N^{(j)}$ are censored.
Further, for each $\theta$, \tabref{tab:censlvl} provides the percent of zeros among the conditional quantiles $q_{1,\theta}(y_1^{(j)} | \vec{x}_1^{(j)}), \dots, q_{N,\theta}(y_N^{(j)} | \vec{x}_N^{(j)})$.
The most challenging cases to model are those with $\theta = 0.05$, where the conditional quantiles are particularly prone to censorship.

\begin{table*}[tb]
    \centering
    \caption[Percent of zero conditional quantiles in each synthetic dataset]{Percent of zero conditional quantiles for the censored observations in each synthetically generated dataset.}
    \label{tab:censlvl}
    \begin{tabular}{l|ccc}
    \toprule
         \thead{Dataset} & $\theta = 0.05$ & $\theta = 0.50$ & $\theta = 0.95$ \\
    \midrule
    Standard Gaussian & $62.7\%$ & $23.9\%$ & $2.0\%$ \\
    Heteroskedastic & $68.0\%$ & $23.9\%$ & $11.4\%$ \\
    Gaussian Mixture & $54.1\%$ & $23.9\%$ & $4.6\%$ \\
    \bottomrule
    \end{tabular}
\end{table*}

For each $j=1,2,3$, we fix train, test and validation sets by randomly partitioning the $j$'th dataset as $62\%:15\%:33\%$, respectively.
To model the $\theta$'th quantile, we use several Neural Networks (NNs), each consisting of a single layer with activation function $\eta$.
Namely, each NN takes as input $\vec{x}$ and outputs either
%\begin{equation}
   %\hat{q}_\theta\left(y | \vec{x}\right) %= \eta\left(\vec{x}\T %\vecbeta\right)\,,
%\end{equation}
\begin{equation}
    %R^2 = 1 - \frac{\sum_{i=1}^N \left(\hat{q}_{\theta, i} - q_{\theta, i}\right)^2}{\sum_{i=1}^N \left(\hat{q}_{\theta, i} - \bar{q}_\theta\right)^2}
    %\,, \qquad
    % \\
    \hat{q}_{i,\theta}\left(y_i | \vec{x}_i\right) = \eta\left(\vec{x}_i\T \vecbeta\right)
    \,,
\end{equation}
for single-output NN, or
\begin{equation}
        \{\hat{q}_{i,\theta,k}\left(y | \vec{x}_i\right)\}_{k=1}^K = \eta\left(\vec{x}_i\T \vecbeta\right)
    \,,
\end{equation}
for multi-output NN, where $\vecbeta$ are trainable parameters (weights). 
We let Multi-CQNN have $K$ times as many trainable parameters as CQNN has (note that there are $K$ independent CQNN's) to keep the number of parameters equal across the different models.
Weights are initialized to $1$, and in each training epoch, the whole train set is processed in a single batch.
Training stops when the validation loss does not improve for $10$ consecutive epochs.

First, as an example of a model that ignores censorship, we use a non-censored linear model, where $\eta$ is the identity function and the loss to be minimized is the tilted-loss without any accounting for censorship.
Namely, for single-output models,
\begin{equation}
    \mathcal{L}(\vecbeta)= \sum_{i=1}^N\rho_\theta\left(\hat{q}_\theta\left(y_i | \vec{x}_i\right) - y_i\right) 
    \,,
    %\rho_\theta \left(\hat{q}_\theta\left(y | \vec{x}\right) - y\right) \,. 
\label{eq:tlloss}
\end{equation}
while for multi-output models,
\begin{equation}
    \mathcal{L}(\vecbeta)= \sum_{k=1}^K\sum_{i=1}^N \bigg(\rho_{\theta,k}\left(\hat{q}_{i,\theta,k}\left(y_i | \vec{x}_i\right) - y_i\right)\bigg)
    \,.
\label{eq:multi_tlloss}
\end{equation}
We refer to these models as the Quantile Neural Network. (\deeplqr\,and \deeplqrmult).
Thereafter, we turn to censorship-aware models, where $\eta$ is the identity function, so that similarly to~\cite{yu2007bayesian}, the single- and multi-output models are linear (\deeplinear\,and \deepmulti).
We measure the predictive performance of each NN on the test set against the actual conditional quantiles of $y^*$ in \eqref{eq:q1}, (\ref{eq:q2}) and (\ref{eq:q3}). 
The measures we use are Mean Absolute Error (MAE) and Rooted Mean Squared Error (RMSE). For any dataset, these measures are defined as follows:
\begin{equation}
        \text{MAE} = \frac{1}{N} \sum_{i=1}^N \left| \hat{q}_{i,\theta} - q_{i,\theta} \right| 
    \,, 
\end{equation}
\begin{equation}
 \text{RMSE} = \sqrt{\frac{1}{N} \sum_{i=1}^N \left( \hat{q}_{i, \theta} - q_{i,\theta} \right)^2}
    \,,
\end{equation}
%\begin{align}
%    \text{MAE} &= \frac{1}{N} \sum_{i=1}^N \left| \hat{q}_{i,\theta} - q_{i,\theta} \right| 
%    \,, \qquad
%     \\
%    \text{RMSE} &= \sqrt{\frac{1}{N} \sum_{i=1}^N \left( \hat{q}_{i, \theta} - q_{i,\theta} \right)^2}
%    \,,
%\end{align}
where $\bar{q}_\theta$ is the mean of $q_{\theta, 1}, \dots, q_{\theta, N}$.
Better predictive quality corresponds to $R^2$ closer to $1$ and MAE and RMSE closer to $0$.

The results appear in \tabref{tab:synth}, which shows that the censorship-aware models outperform the censorship-unaware model.
This holds when evaluating the entire test set or its non-censored subset -- where the latent values are revealed. For the most challenging case of $\theta=0.05$, where more than $60\%$ of the observations are censored, Multi-CQNN is the best performing model across all three synthetic datasets. As the number of censored observations decreases ($\theta=0.95$), we see the difference between the censorship-aware and unaware models becomes less pronounced. 

\begin{table*}[tb]
    \centering
    \caption[Average predictive quality over 10 runs of conditional quantiles of synthetic $y^*$]{Predictive quality for conditional quantiles of $y^*$ in the synthetic datasets. Bold numbers indicate best performance on all test data}
    \label{tab:synth}
\resizebox{0.85\textwidth}{!}{
\begin{tabular}{ll|rr|rr|rr}
\toprule
     & & \multicolumn{2}{c}{\thead{Standard Gaussian}} & \multicolumn{2}{c}{\thead{Heteroskedastic}} & \multicolumn{2}{c}{\thead{Gaussian Mixture}} \\
     $\theta$ & \thead{Model}  & MAE & RMSE & MAE & RMSE &  MAE & RMSE \\
\midrule
\multicolumn{8}{c}{\thead{All Test Data}} \\
\midrule
\multirow{4}{*}{0.05} & \deeplqr &  \ms{1.152}{0.1}& \ms{1.407}{0.1}& \ms{1.223}{0.0}& \ms{1.503}{0.0}& \ms{0.943}{0.0}& \ms{1.151}{0.1} \\
     & \deeplqrmult & \ms{1.104}{0.0}& \ms{1.350}{0.0}& \ms{1.215}{0.0}& \ms{1.500}{0.0}& \ms{0.884}{0.0}& \ms{1.078}{0.0} \\
     & \deeplinear &  \ms{0.947}{0.1}& \ms{1.158}{0.1}& \ms{1.218}{0.0}& \ms{1.501}{0.0}&  \ms{0.826}{0.2}& \ms{1.006}{0.2} \\
     & \deepmulti & \msbest{0.808}{0.0} & \msbest{0.987}{0.0}& \msbest{1.214}{0.0}& \msbest{1.500}{0.0}& \msbest{0.513}{0.1}& \msbest{0.625}{0.1} \\
\cline{1-8}
\multirow{4}{*}{0.50} &  \deeplqr & \ms{0.377}{0.0}& \ms{0.463}{0.0}& \ms{0.383}{0.0}& \ms{0.470}{0.0}& \ms{0.364}{0.0}& \ms{0.445}{0.0} \\
     & \deeplqrmult& \ms{0.378}{0.0}& \ms{0.465}{0.0}&  \ms{0.384}{0.0}& \ms{0.471}{0.0}& \ms{0.364}{0.0}& \ms{0.445}{0.0} \\
     & \deeplinear & \ms{0.163}{0.0}& \ms{0.199}{0.0}&\msbest{0.138}{0.0}& \msbest{0.167}{0.0} & \msbest{0.168}{0.0}& \msbest{0.199}{0.0} \\
     & \deepmulti & \msbest{0.162}{0.0}& \msbest{0.198}{0.0}& \ms{0.139}{0.0}& \ms{0.169}{0.0}& \ms{0.176}{0.0}& \ms{0.209}{0.0} \\
\cline{1-8}
\multirow{4}{*}{0.95} &  \deeplqr & \ms{0.210}{0.0}& \ms{0.261}{0.0}& \msbest{0.534}{0.0}& \msbest{0.619}{0.0}& \ms{0.171}{0.0}& \ms{0.215}{0.0} \\
     & \deeplqrmult& \ms{0.209}{0.0}& \ms{0.260}{0.0}& \ms{0.886}{0.0}& \ms{1.057}{0.0}& \ms{0.289}{0.0}& \ms{0.340}{0.0} \\
     & \deeplinear & \msbest{0.149}{0.0}& \msbest{0.181}{0.0}& \ms{0.635}{0.0}& \ms{0.830}{0.0}& \msbest{0.104}{0.0}& \msbest{0.126}{0.0} \\
     & \deepmulti & \ms{0.156}{0.0}& \ms{0.189}{0.0}& \ms{0.632}{0.0}& \ms{0.826}{0.0}& \ms{0.109}{0.0}& \ms{0.134}{0.0} \\
\midrule
\multicolumn{8}{c}{\thead{Only Non-Censored}} \\
\midrule
\multirow{4}{*}{0.05} & \deeplqr & \ms{0.743}{0.1}& \ms{0.931}{0.1}& \ms{0.507}{0.0}& \ms{0.582}{0.0}& \ms{0.694}{0.1}& \ms{0.881}{0.1} \\
     & \deeplqrmult & \ms{0.643}{0.0}& \ms{0.827}{0.0}& \ms{0.453}{0.0}& \ms{0.534}{0.0}& \ms{0.611}{0.0}& \ms{0.782}{0.0} \\
     & \deeplinear & \ms{0.444}{0.1}& \ms{0.585}{0.1}& \ms{0.478}{0.0}& \ms{0.556}{0.0}& \ms{0.626}{0.2}& \ms{0.784}{0.2} \\
     & \deepmulti & \msbest{0.264}{0.0}& \msbest{0.342}{0.0}& \msbest{0.453}{0.0}& \msbest{0.534}{0.0}& \msbest{0.260}{0.1}& \msbest{0.335}{0.1} \\
\cline{1-8}
\multirow{4}{*}{0.50} &  \deeplqr & \ms{0.258}{0.0}& \ms{0.306}{0.0}& \ms{0.274}{0.0}& \ms{0.331}{0.0}& \ms{0.257}{0.0}& \ms{0.304}{0.0} \\
     & \deeplqrmult & \ms{0.259}{0.0}& \ms{0.306}{0.0}& \ms{0.275}{0.0}& \ms{0.332}{0.0}& \ms{0.257}{0.0}& \ms{0.303}{0.0} \\
     & \deeplinear & \ms{0.111}{0.0}& \msequal{0.132}{0.0}& \msbest{0.098}{0.0}& \msbest{0.117}{0.0}& \msbest{0.122}{0.0}& \msbest{0.141}{0.0}\\
     & \deepmulti & \msbest{0.111}{0.0}& \msbest{0.132}{0.0}& \ms{0.100}{0.0}& \ms{0.119}{0.0}& \ms{0.127}{0.0}& \ms{0.147}{0.0} \\
\cline{1-8}
\multirow{4}{*}{0.95} &  \deeplqr & \ms{0.206}{0.0}& \ms{0.255}{0.0}& \msbest{0.539}{0.0}& \msbest{0.623}{0.0}& \ms{0.167}{0.0}& \ms{0.210}{0.0} \\
     & \deeplqrmult & \ms{0.205}{0.0}& \ms{0.255}{0.0}& \ms{0.828}{0.0}& \ms{1.001}{0.0}& \ms{0.272}{0.0}& \ms{0.314}{0.0} \\
     & \deeplinear & \msbest{0.150}{0.0}& \msbest{0.183}{0.0}& \ms{0.647}{0.0}& \ms{0.840}{0.0}& \msbest{0.103}{0.0}& \msbest{0.125}{0.0} \\
     & \deepmulti &\ms{0.157}{0.0}& \ms{0.190}{0.0}& \ms{0.644}{0.0}& \ms{0.836}{0.0}& \ms{0.108}{0.0}& \ms{0.134}{0.0} \\
\bottomrule
\end{tabular}
    }

\vspace{-0.5cm}
\end{table*}

\subsection{Parametric vs. Non-Parametric Censored Quantile Regression}

Since its introduction by \cite{tobin1958estimation}, the Tobit model has become a cornerstone of parametric censored modeling. 
Tobit assumes that the latent variable depends on covariates linearly with Gaussian white noise, and is censored at a given fixed threshold.
Hence in Tobit, the latent quantiles for the $i$'th observation are given by the parametric distribution
\begin{equation}
    \mathcal{N} \left( \vec{x}_i\T \vecbeta \,, \sigma^2 \right)
    \,,
\end{equation}
where $\vec{x}_i$ are covariates, $\vecbeta$ are linear coefficients to be estimated, and $\sigma$ is standard deviation, either given or to be estimated too.
Further, the Tobit likelihood for right censorship is
\begin{equation}
    \prod_{i=1}^N 
    \Bigg \{
    \frac{1}{\sigma} \varphi\left(\frac{y_i - \vec{x}_i\T \vecbeta}{\sigma}\right) 
    \Bigg \}^{(1 - \lbl_i)} 
    \Bigg \{
    1 - \Phi\left(\frac{y_i - \vec{x}_i\T \vecbeta}{\sigma}\right)
    \Bigg \}^{\lbl_i}
    \,,
\label{eq:tobitlik}
\end{equation}
where $\varphi$ is the Probability Density Function (PDF) of $\mathcal{N}(0, 1)$, $\Phi$ is its Cumulative Distribution Function (CDF), and for a given fixed threshold $\tau_i$:
\begin{equation}
\lbl_i = 
\begin{cases} 
0 \,, & y_i < \tau_i \\
1 \,, & y_i = \tau_i \\
\end{cases}
\,.
\end{equation}
The Tobit negative log-likelihood is then
\begin{multline}
\label{eq:tobit_fit}
    \mathcal{L}(\vecbeta)=-\sum_{i=1}^N \lbl_i\log \left(
 \varphi\left(\frac{y_i - \vec{x}_i\T \vecbeta}{\sigma}\right) \right) \\
     +(1-\lbl_i)\log\left(1-\Phi\left(\frac{y_i - \vec{x}_i\T \vecbeta}{\sigma}\right)\right)
    \,,
\end{multline}
which we use as a loss function and minimise as described for the previous models.

Let us now compare Tobit parametric modeling to non-parametric CQNN, using the same synthetic baseline datasets as above.
For both modeling methods, we use an NN with a single linear neuron, which we fit similarly to \secref{sec:dcqr-exp-nocens}.
When fitting the Tobit model, we fix $\sigma = 1$ and use the NLL of \eqref{eq:tobit_fit} as the loss function, whereas when fitting CQR for $\theta = 0.05, 0.95$, we use the NLL of \eqref{eq:ourlik} or \eqref{eq:multiourlik} as the loss function. The quantiles from the Tobit model will then correspond to quantiles in the Gaussian distribution $\mathcal{N}\left( \vec{x}_i\T \vecbeta \,, \sigma^2 \right)$

Finally, we evaluate the performance via the Tilted Loss~\eqref{eq:tl} as well as two common measures of Quantile Regression, Interval Coverage Percentage (ICP) and Mean Interval Length (MIL): 
\begin{align}
    %\text{Interval Coverage Percentage (ICP)    } &= \text{fraction of $y^*$ within the estimated prediction interval.} \label{eq:icp}\\
    %\text{Mean Interval Length (MIL)} &= \text{mean length of the estimated predictions interval.} \label{eq:mil}
    \text{ICP} &= \frac{1}{N}\sum_{i=1}^N\begin{cases}
    1 & \text{ if } \hat{q}_{i,\theta} \leq y_i \leq \hat{q}_{i,\theta'}   \\
    0 & \text{otherwise}
    \end{cases} \label{eq:icp} \\
    \text{MIL} &= \frac{1}{N}\sum_{i=0}^N(|\hat{q}_{i,\theta}-\hat{q}_{i,\theta'}|) \label{eq:mil}
\end{align}
where $\hat{q}_{i,\theta}$ is the predicted $\theta$ quantile for observation $i$ and $\theta \leq \theta'$, so $\theta'$ is a higher quantile than $\theta$. 
For both measures, we define the prediction interval as the interval between the  $0.05$'th-quantile and the $0.95$'th-quantile. The ICP should be close to $0.95 - 0.05 = 0.9$, while MIL should be as small as possible.
Among models with same ICP, we thus prefer the one that yields the lowest MIL.

\tabref{tab:qrvstobit} summarizes the performance of Tobit vs. the QR models, i.e., CQNN and Multi-CQN.
As expected, Tobit performs overall best on the synthetic dataset with Standard Gaussian noise, which most closely matches its modeling assumptions.
When evaluated on all test observations, Tobit outperforms QR by obtaining an ICP closer to the desired $0.9$ for each synthetic dataset.
However, when evaluated on just the non-censored test observations (approx. $30\%$ of each dataset), where the actual values are reliably known, QR outperforms Tobit while maintaining ICP close to $0.9$. A particular limitation of the Tobit model is the relatively high MIL (which is constant in Tobit with fixed variance) compared to the CQNN models. The results thus suggest that Multi-CQNN tends to yield flatter distributions (higher MIL) that better approximate the latent distribution of non-censored observations.

\begin{table*}[tb]
    \centering
    \caption[Non-parametric QR vs. parametric Tobit on synthetic datasets]{Non-parametric QR vs. parametric Tobit on synthetic datasets. Lowest Tilted Loss is highlighted in bold.}
\label{tab:qrvstobit}
    \resizebox{0.85\textwidth}{!}{
    \begin{tabular}{ll|rrr|rrr}
    \toprule
    & & \multicolumn{3}{c}{All Test Data} & \multicolumn{3}{c}{Only Non-Censored} \\ 
    \thead{Dataset} & \thead{Model} & \thead{ICP} & \thead{MIL} & \thead{Tilted Loss} & \thead{ICP} & \thead{MIL} & \thead{Tilted Loss} \\
    \midrule
    \multirow{3}{*}{Standard Gaussian} & Tobit &\msbest{0.91}{0.0}& \msbest{3.29}{0.0}& \msbest{65.81}{0.2} & \ms{0.91}{0.0}& \ms{3.29}{0.0}& \ms{47.13}{0.1}  \\

    & \deeplinear & \ms{0.68}{0.0}& \ms{2.50}{0.0}& \ms{107.95}{8.7} & \ms{0.89}{0.0}& \ms{2.87}{0.1}& \ms{45.46}{0.6}\\
    & \deepmulti & \ms{0.72}{0.0}& \ms{2.49}{0.0}& \ms{93.10}{1.7} & \msbest{0.89}{0.0}& \msbest{2.77}{0.0}& \msbest{44.65}{0.1} \\
    \midrule
    \multirow{3}{*}{Heteroskedastic} & Tobit & \msbest{0.85}{0.0}& \msbest{3.29}{0.0}& \msbest{113.12}{0.1} & \ms{0.84}{0.0}& \ms{3.29}{0.0}& \ms{78.67}{0.2} \\
    & \deeplinear & \ms{0.61}{0.0}& \ms{2.73}{0.0}& \ms{170.29}{0.6} & \ms{0.89}{0.0}& \ms{3.32}{0.0}& \ms{66.28}{0.7}\\
    & \deepmulti & \ms{0.61}{0.0}& \ms{2.72}{0.0}& \ms{170.32}{0.4} & \msbest{0.89}{0.0}& \msbest{3.32}{0.0}& \msbest{66.18}{0.5}\\
    \midrule
    \multirow{3}{*}{Gaussian Mixture} & Tobit & \msbest{0.94}{0.0}& \msbest{3.29}{0.0}& \msbest{61.26}{0.1} & \ms{0.94}{0.0}& \ms{3.29}{0.0}& \ms{44.09}{0.2} \\

    & \deeplinear & \ms{0.70}{0.0}& \ms{2.38}{0.1}& \ms{102.53}{16.3} & \ms{0.90}{0.0}& \ms{2.77}{0.1}& \ms{41.56}{0.7} \\
    & \deepmulti & \ms{0.75}{0.0}& \ms{2.36}{0.0}& \ms{76.89}{6.8} & \msbest{0.88}{0.0}& \msbest{2.61}{0.0}& \msbest{40.93}{0.1} \\
    \bottomrule
    \end{tabular}
    }
\end{table*}

\subsection{Single-output vs. Multi-output Censored Quantile Regression} \label{sec:dcqr-exp-crossing}
We now turn to the evaluation of the quantile crossing problem for the CQNN models. For this, we fit multiple single-output CQNN and one Multi-CQNN to estimate the deciles of the three different synthetic datasets.
For all the estimated deciles it should hold that $\hat{q}_{\theta, 1} \leq \hat{q}_{\theta, 2} \leq \cdots \leq \hat{q}_{\theta, K}$ assuming that $\hat{q}_{\theta,j+1}$ is a larger decile than $\hat{q}_{\theta,j}, \forall j \in 1,\dots,K-1$. 
We evaluate violations of this order via two measures of quantile crossings:
\begin{align}
    \text{Total number of crosses} &=\sum_{i=1}^N\sum_{k=1}^{K-1} \left(q_{i,\theta,k} \geq  q_{i,\theta,k+1}\right), \label{eq:toc}
    \\
    \text{Crossing Loss (CL)} &=\sum_{i=1}^N\sum_{k=1}^{K-1}\max\left(0,q_{i,\theta,k}-q_{i,\theta,k+1}\right)\,.
     \label{eq:CL}
\end{align}
Lower values of these measures correspond to fewer quantile crossings, with zero corresponding to the best case of no crossings at all. 

\tabref{tab:crosses} summarize the quantile crossing performance of the models. We observe that the total number of crossing is substantially less for the Multi-CQNN compared to the single output CQNN. This is consistent across all the three synthetic datasets, both for the censored and non-censored observations, with less computational complexity than $K$ single-output NN's.
In conclusion, our experiments on the synthetic datasets have shown that the proposed Multi-CQNN model outperforms single-output CQNN in terms of quantile crossing, ICP and MIL.

\begin{table*}[tb]
    \centering
    \caption{Quantile crossings in single- and multi-output CQNN.}
    \label{tab:crosses}
    \resizebox{0.85\textwidth}{!}{
    \begin{tabular}{ll|rr|rr}
    \toprule
    & & \multicolumn{2}{c}{All Test Data} & \multicolumn{2}{c}{Only Non-Censored} \\ 
    \thead{Dataset} & \thead{Model} & \thead{Total number of crosses} & \thead{CL} & \thead{Total number of crosses} & \thead{CL} \\
    %Dataset & Model & Cross loss & Total number of crosses  \\
    
    \midrule
    \multirow{2}{*}{Standard Gaussian}
    % & \deeplqr & \ms{53.70}{21.3}& \ms{0.03}{0.0}& \ms{7.30}{5.4}& \ms{0.63}{0.6} \\
    % & \deeplqrmult & \ms{16.40}{2.5}& \ms{0.00}{0.0}& \ms{0.00}{0.0}& \ms{0.00}{0.0} \\
     & \deeplinear & \ms{67.40}{45.0}& \ms{0.04}{0.0}& \ms{16.90}{12.8}& \ms{2.18}{1.9} \\
     & \deepmulti & \msbest{7.60}{0.5}& \msbest{0.00}{0.0}& \msbest{0.00}{0.0}& \msbest{0.00}{0.0} \\ 
    \midrule
    \multirow{2}{*}{Heteroskedastic}
    % & \deeplqr & \ms{194.90}{59.5}& \ms{0.05}{0.0}& \ms{50.70}{30.7}& \ms{2.35}{0.9} \\
    %& \deeplqrmult & \ms{167.20}{33.5}& \ms{0.04}{0.0}& \ms{34.90}{17.7}& \ms{0.01}{0.0} \\ 
     & \deeplinear & \ms{399.50}{14.2}& \ms{0.19}{0.0}& \ms{90.30}{4.3}& \ms{7.82}{1.5}\\ 
     & \deepmulti & \msbest{391.10}{1.9}& \msbest{0.17}{0.0}& \msbest{84.10}{0.7}& \msbest{6.86}{0.0}\\
    \midrule
    \multirow{2}{*}{Gaussian Mixture} 
    %& \deeplqr& \ms{55.00}{37.7}& \ms{0.02}{0.0}& \ms{8.40}{6.2}& \ms{0.69}{0.6} \\
     %& \deeplqrmult& \ms{13.70}{0.7}& \ms{0.00}{0.0}& \ms{0.10}{0.3}& \ms{0.00}{0.0} \\
     & \deeplinear & \ms{51.90}{20.5}& \ms{0.01}{0.0}& \ms{7.30}{5.3}& \ms{0.30}{0.3} \\
     & \deepmulti & \msbest{20.00}{1.3}& \msbest{0.00}{0.0}& \msbest{2.10}{0.3}& \msbest{0.07}{0.0} \\
    \bottomrule
    \end{tabular}
    }
\end{table*}

\section{Experiments for Estimating Latent Mobility Demand} \label{sec:dcqr-exp-real}
In this Section, we apply our Multi-Output Censored Quantile Regression Neural Network (Multi-CQNN) to real-world data from shared mobility services.
Contrary to the synthetic datasets in the previous Section, real-world datasets do not feature the latent variable.
Hence, similarly to \cite{gammelli2020estimating}, we treat the available data as $y^*$ and manually censor it per various censorship schemes.
We then fit CQNN and Multi-CQNN models for $\theta=0.05,0.95$ and evaluate them via ICP \eqref{eq:icp} and MIL \eqref{eq:mil}.

First, we use a censorship-unaware NN with a single linear unit, which we train to minimize plain Tilted Loss \eqref{eq:tlloss} (denoted QNN).
Then, we equip the same architecture with censorship-awareness, using \eqref{eq:ourlik} as loss (CQNN).
To show the versatility of our proposed approach, we experiment with the addition of Long Short-Term Memory (LSTM)~\cite{hochreiter1997long} to the CQNN (CQNN+LSTM), which we extend to Multi-CQNN. We chose he LSTM architecture based on its extensive use in the transportation domain~\cite{himanen2019neural}, and note that our approach can similarly be used with any desired architecture.

\subsection{Bike-sharing Data} \label{sec:bike}

\begin{figure}[tb]
    \centering
    \includegraphics[width=0.45\textwidth]{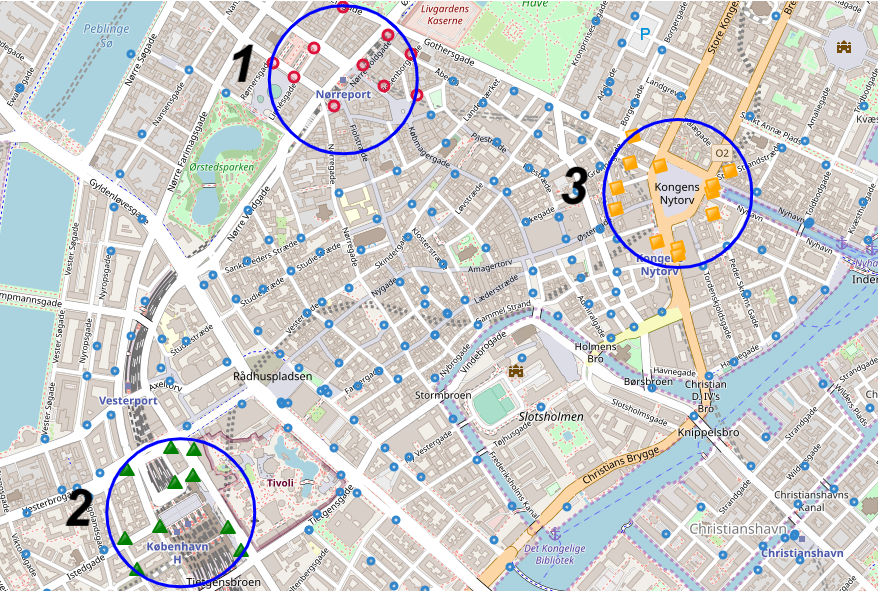}
    \caption[Bikes-haring hubs and aggregated superhubs]{Bike-sharing hubs and aggregated superhubs, as in \cite{gammelli2020estimating}: 1) N{\o}rreport, 2) Kgs. Nytorv, 3) K{\o}benhavn H.}
\end{figure}

%\begin{figure*}[!bt]
%\centering
%\includegraphics[width=0.6\textwidth]{icp_mean_all.png}
%\label{fig:fig_first_case}
%\end{figure*}

\begin{figure*}[!tb]
    \centering
    \includegraphics[width=0.8\textwidth]{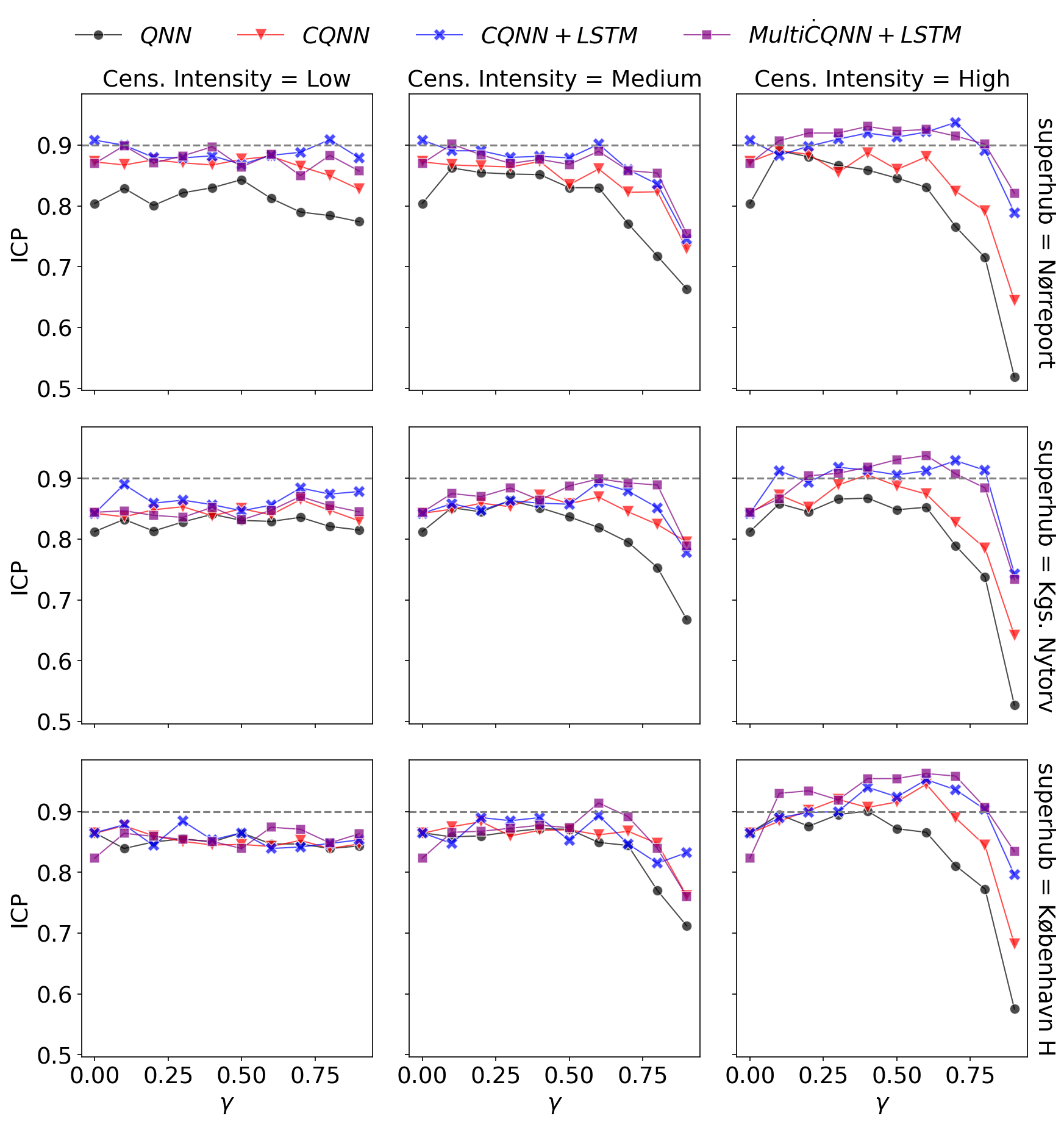}
    \caption[Mean ICP for bike-sharing data, evaluated on all test observations. $0.9$ ICP is marked with horizontal gray line.]{Mean ICP for bike-sharing data, evaluated on all test observations. $0.9$ ICP is marked with horizontal gray line.}
    \label{fig:icp_mean}
\end{figure*}

\begin{figure*}[!bt]
    \centering
    \includegraphics[width=0.8\textwidth]{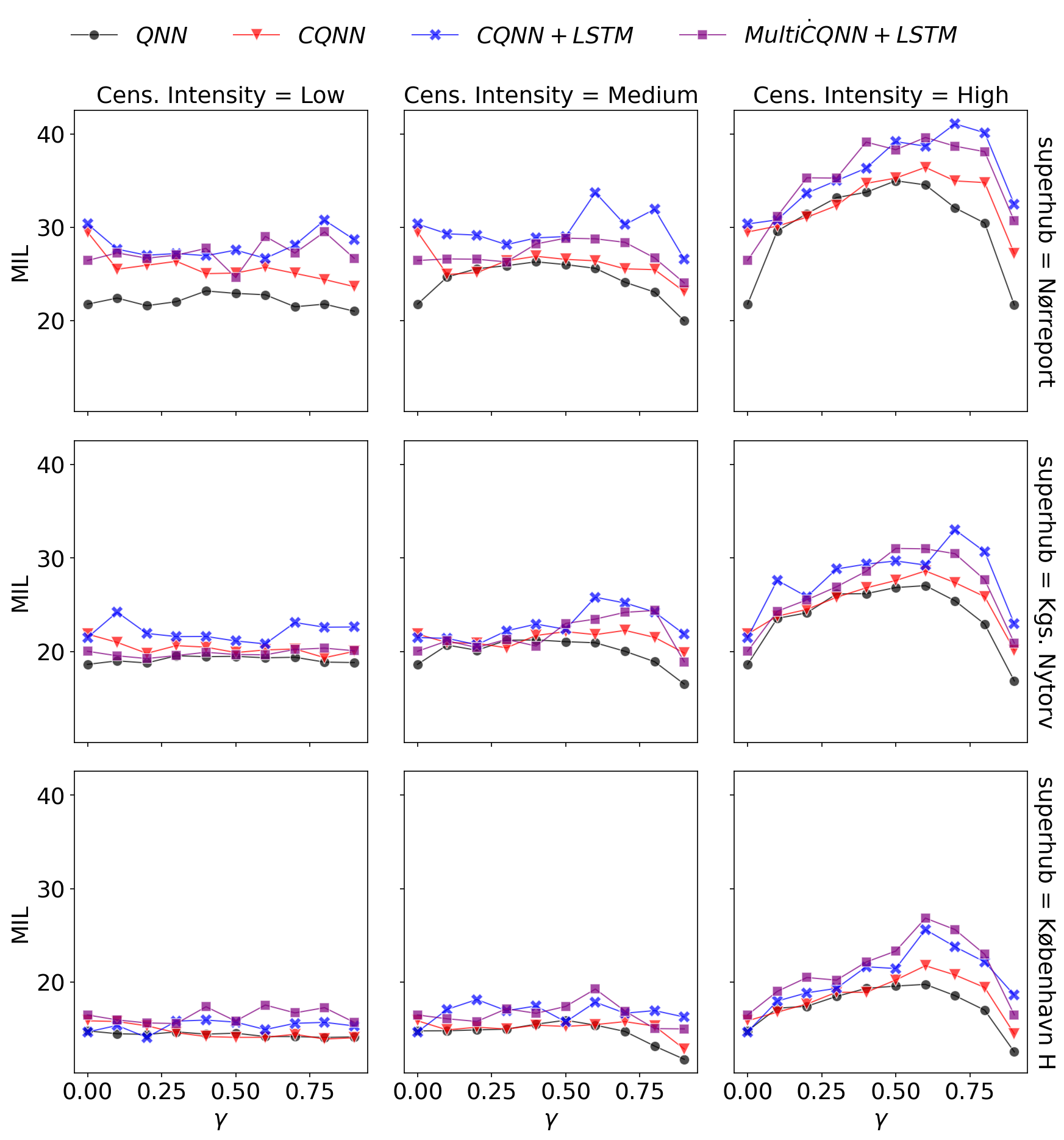}
    \caption[MIL (b) for bike-sharing data, evaluated on all test observations.]{MIL for bike-sharing data, evaluated on all test observations.}
    \label{fig:mil_mean}
\end{figure*}

\iffalse
% used to be 3in
%\begin{figure*}[!bt]
%\centering
%\subfloat[Mean ICP]{\includegraphics[width=4in]{icp_mean_all.png}%
%\hfil
%\subfloat[Mean MIL]{\includegraphics[width=4in]{mil_mean_all}%
%\label{fig_second_case}}
%\caption{Mean ICP (a) and MIL (b) for bike-sharing data, evaluated on all test observations. $0.9$ ICP is marked with horizontal gray line (a).}
%\label{fig:resbike}
%\end{figure*}

\begin{figure*}[tb]
    \centering
    \begin{subfigure}{\textwidth}
    \centering
    \includegraphics[width=0.45\textwidth]{icp_mean_all.png}
    \hfill
    \includegraphics[width=0.45\textwidth]{mil_mean_all.png}
    \caption[Bike-sharing results on non-censored test observations]{Mean ICP (left) and MIL (right) for bike-sharing data, evaluated on all test observations. $0.9$ ICP is marked with horizontal gray line (left).}
    \label{fig:bikenc}
    \end{subfigure}
    %\caption{Results for bike-sharing data.}
    \label{fig:resbike}
\end{figure*}
\fi

The first real-world dataset is from Donkey Republic, a bike-sharing service provider in the Copenhagen metropolitan area in Denmark.
The data consists of pickups and returns of bicycles in predefined hubs (in total 32 hubs), from 1 March 2018 until 14 March 2019, which we aggregate spatially into 3  ``superhubs'' and temporally by no. daily pickups daily, as in \cite{gammelli2020estimating}. The superhubs are chosen based on distance from main tourist attractions and the central train station.
Superhubs rarely run out of bicycles at any moment; hence this data represents actual demand quite well.

For this dataset, we experiment with partial censorship of the daily demand. This scenario occurs when the supply of bikes is lower than the actual demand for bicycles, which corresponds to lost opportunities for the bike-sharing provider.

We censor the data as follows:
\begin{enumerate}
    \item Randomly select a $\gamma$ portion of all $y^*_i$.
    \item For each selected $y^*_i$, independently sample 
    \begin{equation}
    \delta_i \sim \mathcal{U}\left[c_1, c_2\right]
    \,,
    \end{equation}
    and let
    \begin{equation}
    y_i = (1 - \delta_i)y^*_i
    \,.
    \end{equation}
\end{enumerate}
Our experiments use $\gamma = 0.0, 0.1, \dots, 0.9$ and $\left(c_1, c_2\right) =$ $\left(0.01, 0.33\right)$, $\left(0.34, 0.66\right)$, $\left(0.67, 0.99\right)$.
We define these values of $(c_1, c_2)$ as Low, Medium, and High censorship intensity.
For each $\gamma, c_1, c_2$, we independently censor the data $10$ times to obtain differently censored datasets $B_1, \dots, B_{10}$, and we partition each $B_j$ consecutively into train, validation and test sets with equal proportions.
We then fit each NN model independently for $10$ random initialisations of weights, drawn independently from $\mathcal{N}(0, 1)$, where we use the $7$ previous lags of observations as explanatory variables. We define the censorship thresholds for non-censored observations as:
\begin{equation}
    \tau_i^{(j)} = y_i^{(j)} \times \frac{\text{mean of train $y^*$}}{\text{mean of train $y^{(j)}$}}
    \,.
\end{equation}
An observation is then censored if and only if it is above the threshold.
After fitting, we evaluate ICP and MIL for each $\gamma, c_1, c_2$ as follows.
We noticed that some experiments would result in the MIL being unreasonable high and decided to exclude these experiments in the results. Hence, for each of $B_1, \dots, B_{10}$, we consider only initialisations that yield reasonable validation MIL as:

\begin{equation}
    \frac{\text{validation MIL}}{\text{mean of train $y^{(j)}$}} \leq 2
    \,.
\end{equation}
We then select the initialisation that yields validation ICP closest to $0.9$.
Finally, we average the test ICP and test MIL over the $10$ selected initialisations.
We summarize the results for the bike-sharing data experiments in \figref{fig:icp_mean} and \figref{fig:mil_mean} for the entire test set.
In each Figure, rows range over superhubs, columns run over the censorship intensity, and each horizontal axis ranges over $\gamma$.

We see that the worst-performing model is the censorship unaware. The ICP (\figref{fig:icp_mean}) for the unaware model is substantially lower than the censorship aware models, and this difference becomes more prominent as the number of censored observations increases.
We note that, as we increase complexity in the architecture, we find higher MIL (\figref{fig:mil_mean}) and generally better ICP.
The censorship-unaware model often yields the worst ICP. However, as found in~\cite{gammelli2020estimating}, its ICP is occasionally better than that of the censorship-aware models, when relatively few observations are censored ($\gamma \leq 0.2$).
We also see that among CQNN models, the LSTM-based model often yields better ICP than the purely linear model. 
We note that the Multi-CQNN tends to have higher ICP than single-output CQNN, with a larger MIL as a trade-off between the ICP and MIL. A wider MIL may be a desired property to express uncertainty in the outputs from a conservative perspective.

\subsection{Shared Electric Vehicles Data} \label{sec:ev}

\begin{figure}[!tb]
    \centering
    \includegraphics[width=0.45\textwidth]{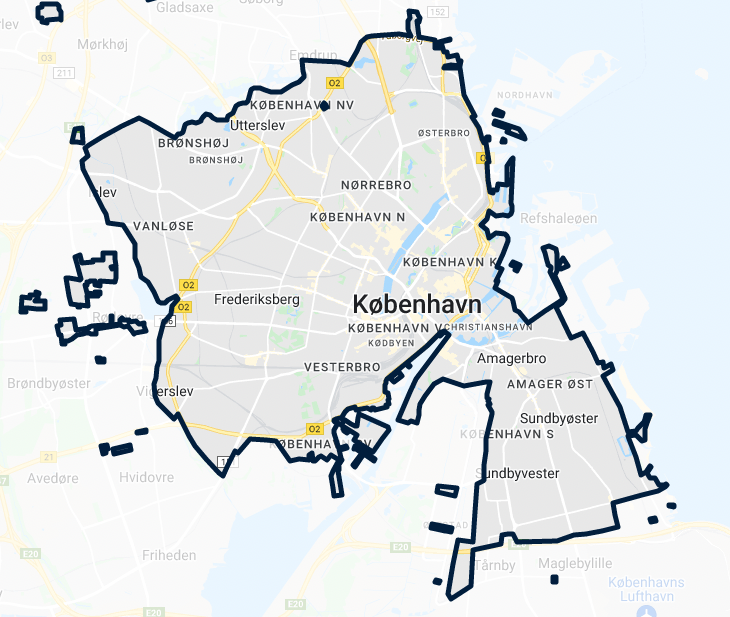}
    \caption[Overview of Copenhagen metropolitan area, where shared EVs can be picked up and returned]{Overview of Copenhagen metropolitan area, where shared EVs can be picked up and returned~\cite{sharenow}.}
    \label{fig:sharenow_overview}
\end{figure}

The second real-world dataset comes from Share Now, a shared Electric Vehicles (EVs) service operator in the Copenhagen metropolitan area too. 
Users can pick up designated EVs from any location in the metropolitan area and return them to any parking spot within the region. In addition, there are small satellite locations where cars be picked up and returned~(\figref{fig:sharenow_overview}).
This dataset, which we denote as \evdata, consists of \SI{2.6}{\million} trips from 2016 to 2019, where each trip record contains the endpoints, driver ID and vehicle ID.

For this dataset, we experiment with complete censorship of daily demand of EV mobility, wherein all observations are censored. A scenario where this occurs is when multiple providers are competing for the same services. For example, one company might only observe the demand from their own fleet of EVs, and therefore all its observations are censored, as some demand is served by the competition.

\begin{figure*}[tb]
    \centering
    \includegraphics[width=\textwidth]{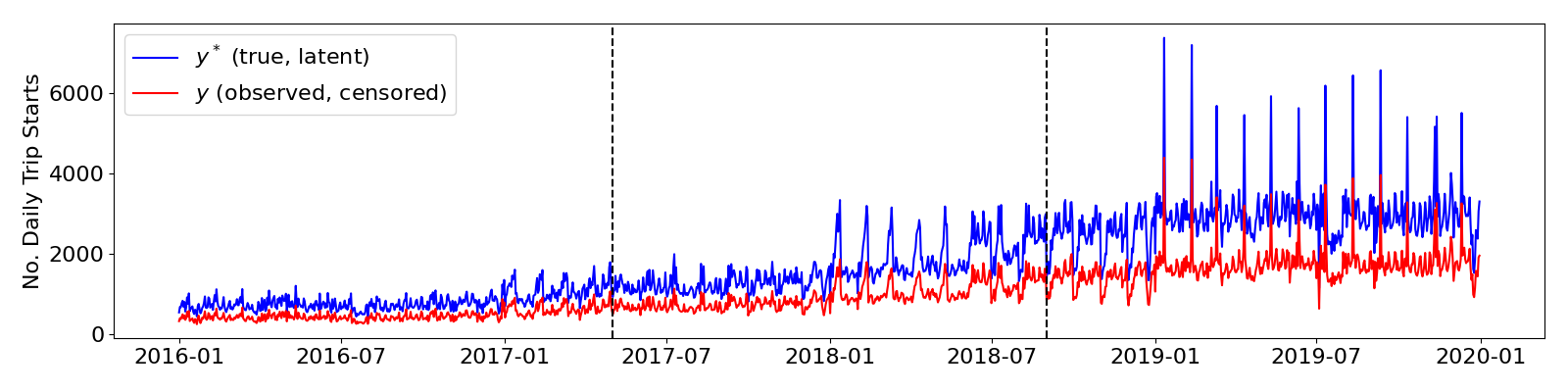}
    \caption{Shared EV data with $40\%$ fleet reduction~($\alpha=0.4$). Dashed lines indicate, from left to right, the train, validation and test set.}
    \label{fig:fleetcens}
\end{figure*}

First, we let $y^*_1, \dots, y^*_N$ be the daily no. trip starts in \evdata. 
For censorship, we use the following scheme:
\begin{enumerate}
    \item Randomly select an $\alpha$ portion of all vehicles in \evdata.
    \item Let \evdatacens\ be \evdata without any trips that involve the selected vehicles.
    \item Let $y_1, \dots, y_N$ be the daily no. trip starts in \evdatacens.
\end{enumerate}
Consequently, every $y_i$ is censored, so that
\begin{equation}
    y_i \approx (1 - \alpha) y^*_i
    \,,
\end{equation}
as illustrated in \figref{fig:fleetcens}.

Since we censor all the observations, there is no need to specify a censoring threshold for them.
For each $\alpha=10\%,20\%,30\%,40\%$, we independently apply the censorship scheme $10$ times.
For each of the $10$ censored datasets thus obtained, we partition into train:validation:test as $1:1:1$ and fit each NN with random initialisation of weights, drawn independently from $\mathcal{N}(0, 1)$.
Finally, we evaluate each NN by averaging its test ICP, MIL, and CL over the $10$ experiments.

The results appear in \tabref{tab:ressharenow}, where we see again 
that the censorship-unaware model mostly yields both the worst ICP and worst MIL.
Among the censorship-aware models, Multi-CQNN with LSTM often yields the best ICP, and otherwise has ICP close to the ICP of the single-output LSTM model.
For censorship-aware models, too, better ICP is accompanied by higher MIL, as in \secref{sec:bike}.
As expected with complete censorship, all models deteriorate rapidly as $\alpha$ increases, yielding ICP far below $0.9$ for $\alpha = 0.4$, where our experiments thus stop. Perhaps this deterioration would be less pronounced with larger datasets for higher levels of $\alpha$'s.

\begin{table*}[tb]
    \centering
    \caption[Results of experiments with shared EV data]{Results of experiments with shared EV data, as average $\pm$ standard deviation. We highlight the best-performing model based on ICP in bold. The lowest MIL breaks ties.}
    \label{tab:ressharenow}
    \begin{adjustbox}{max width=0.99\textwidth}
    \setlength\tabcolsep{3pt}
    \begin{tabular}{cc|rrrrrr}
    \toprule
     Metric & $\alpha$-Level & \thead{\deeplqr} & \thead{CQNN} & \thead{Multi-CQNN} & \thead{CQNN+LSTM} & \thead{Multi-CQNN+LSTM} \\
\midrule
\multirow{4}{*}{ICP} & $0.1$ & \ms{0.870}{0.0} & \ms{0.911}{0.0} & \ms{0.913}{0.0} & \msbest{0.894}{0.0} & \ms{0.906}{0.0}\\
& $0.2$ & \ms{0.659}{0.0} & \ms{0.858}{0.0} & \ms{0.854}{0.0} & \ms{0.876}{0.1} & \msbest{0.878}{0.0}\\
& $0.3$ &  \ms{0.341}{0.1} & \ms{0.683}{0.1} & \ms{0.658}{0.1} & \ms{0.681}{0.2} & \msbest{0.759}{0.1}\\
& $0.4$ & \ms{0.106}{0.0} & \ms{0.282}{0.1} &  \ms{0.267}{0.1} & \ms{0.313}{0.2} & \msbest{0.411}{0.2}\\
\midrule
\multirow{4}{*}{MIL} & $0.1$ & \ms{1680}{123} & \ms{2015}{71} & \ms{2075}{113} & \msbest{2081}{191}& \ms{2567}{364}\\
& $0.2$ &  \ms{1442}{75} &  \ms{1848}{84} & \ms{1925}{96} & \ms{2504}{929} & \msbest{2173}{286}\\
& $0.3$ &   \ms{1339}{84} & \ms{1641}{109} & \ms{1671}{92} & \ms{1959}{598} & \msbest{1993}{169}\\
& $0.4$ & \ms{1105}{62} & \ms{1406}{48} & \ms{1415}{111} & \ms{1574}{276} & \msbest{1727}{159}\\
\midrule
\multirow{4}{*}{CL} & $0.1$ & \ms{0.000}{0.0} & \ms{0.000}{0.0} & \ms{0.000}{0.0} & \msbest{0.002}{0.0} & \ms{0.000}{0.0}\\
& $0.2$ & \ms{0.000}{0.0} & \ms{0.000}{0.0} & \ms{0.000}{0.0} & \ms{0.002}{0.0} & \msbest{0.000}{0.0}\\
& $0.3$ &  \ms{0.000}{0.0} & \ms{0.000}{0.0} & \ms{0.000}{0.0} & \ms{0.000}{0.0} & \msbest{0.000}{0.0}\\
& $0.4$ & \ms{0.000}{0.0} & \ms{0.000}{0.0} & \ms{0.000}{0.0} & \ms{0.000}{0.0} & \msbest{0.000}{0.0}\\
\bottomrule
\end{tabular}
\end{adjustbox}
\end{table*}

\section{Conclusion} \label{sec:dcqr-conclusion}
In summary, we have addressed the problem of censored mobility demand and proposed to estimate the entire distribution of latent mobility demand via Multi-Output Censored Quantile Regression Neural Networks (Multi-CQNN). Our approach mitigates the problem of censored and uncertain demand estimation, which is vital for mobility services driven by the user demand, in order to plan supply accordingly.

First, we demonstrate the advantages of censorship-aware models on synthetic baseline datasets with various noise distributions, both homoskedastic and heteroskedastic.
We find that CQNN outperforms censorship-unaware QNN on both the entire test set and its non-censored subset, where the actual values are reliably known. 
We also compare Multi-CQNN to the standard Tobit model, which assumes Gaussian white noise, and obtain that Multi-CQNN tends to yield flatter distributions that better approximate the latent uncertainty structure of non-censored observations.
We also show that our proposed multi-output extension to CQNN produces substantially fewer quantile crossings for censored and non-censored observations.

Next, we apply Multi-CQNN to real-world datasets from two shared mobility services -- bike-sharing and shared Electric Vehicles (EVs) -- which we randomly censor either partially or entirely.
For both datasets, more complex CQNN architectures yield higher MIL and generally better ICP. Adding a Long Short-Term Memory (LSTM) often leads to the best performance, and censorship-unaware QNN often produces the worst ICP. We observe that Multi-CQNN is performing on par with the single-output models while requiring less computational resources. In all experiments, we observe that the Multi-CQNN model tends to outperform the CQNN as censorship intensifies.

The experiments on synthetic and real-world datasets thus lead to similar conclusions about the effectiveness of the Multi-CQNN for censored regression, which further emphasizes the importance of accounting for censorship when modeling mobility demand.
For future work, we plan to take advantage of possible Spatio-temporal correlations in the datasets, e.g., using Convolutional Neural Networks as in \cite{chu2019deep} or Graph Neural Networks~\cite{Zhao2020tgcn}, as well as compare Multi-CQNN to Censored Gaussian Processes~\cite{gammelli2020estimating}. In addition, we propose to explore the impact of censored regression in the operation of mobility services which is driven by demand~\cite{gamelli2022predictive}.

\section{Acknowledgements}
The research leading to these results has received funding from the Independent Research Fund Denmark (Danmarks Frie Forskningsfond) under the grant no. 0217-00065B.

\bibliographystyle{IEEEtran}
\bibliography{ref}

\section{Biography Section}
\begin{IEEEbiography}[{\includegraphics[width=1in,height=1.25in,clip,keepaspectratio]{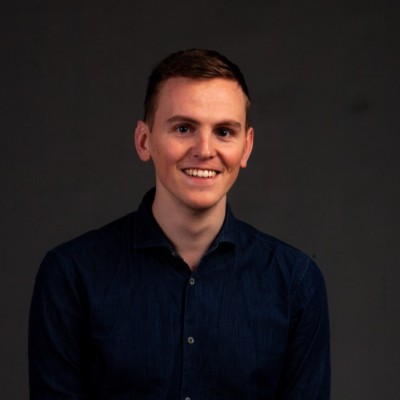}}]{Frederik Boe Hüttel} is is currently pursuing the Ph.D.
degree in Machine Learning for Smart Mobility with the Technical University of Denmark (DTU). His main research interests include machine learning models, intelligent transportation systems, and demand modeling.
\end{IEEEbiography}

\begin{IEEEbiography}[{\includegraphics[width=1in,height=1.25in,clip,keepaspectratio]{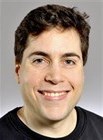}}]{Inon Peled} is a recent Ph.D. graduate from the Techical University of Denmark. He received a Ph.D. Dergree in Machine Learning 2020 from the Technical University of Denmark (DTU). His research interests include Machine Learning under uncertainty, Big Data, modeling and Prediction, and Autonomous and Green Mobility.
\end{IEEEbiography}

\begin{IEEEbiography}[{\includegraphics[width=1in,height=1.25in,clip,keepaspectratio]{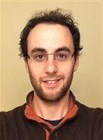}}]{Filipe Rodrigues} is Associate Professor at the Technical University of Denmark (DTU), where he is working on machine learning models for understanding urban mobility and the behaviour of crowds, with emphasis on the effect of special events in mobility and transportation systems. He received a Ph.D. degree in Information Science and Technology from University of Coimbra, Portugal, where he developed probabilistic models for learning from crowdsourced and noisy data. His research interests include machine learning, probabilistic graphical models, natural language processing, intelligent transportation systems and urban mobility.
\end{IEEEbiography}

\begin{IEEEbiography}[{\includegraphics[width=1in,height=1.25in,clip,keepaspectratio]{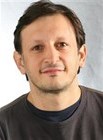}}]{Francisco C. Pereira} is Full Professor at the Technical University of Denmark (DTU), where he
leads the Smart Mobility research group. His main research focus is on applying machine learning and
pattern recognition to the context of transportation systems with the purpose of understanding and predicting mobility behavior, and modeling and optimizing the transportation system as a whole. He has Masters (2000) and Ph.D. (2005) degrees in Computer Science from University of Coimbra, and has
authored/co-authored over 70 journal and conference papers in areas such as pattern recognition, transportation, knowledge based systems and cognitive science. Francisco was previously a Research Scientist at MIT and Assistant Professor in University of Coimbra. He was awarded several prestigious prizes, including an IEEE Achievements award, in 2009,the Singapore GYSS Challenge in 2013, and the Pyke Johnson award from Transportation Research Board, in 2015.
\end{IEEEbiography}

\vfill
\end{document}